\documentclass[sigconf]{acmart}
\usepackage{booktabs} 
\usepackage{soul}
\usepackage{graphicx}
\usepackage{subfigure}
\usepackage{colortbl}

\newcommand{\pop}{\boldsymbol{P}}
\newcommand{\pare}{\boldsymbol{P}_{par}}
\newcommand{\offs}{\boldsymbol{P}_{off}}
\newcommand{\sur}{\boldsymbol{P}_{sur}}
\newcommand{\cand}{\boldsymbol{P}_{can}}
\newcommand{\arc}{\boldsymbol{A}}
\newcommand{\score}{\boldsymbol{S}}

\begin{document}
\title{SHX: Search History Driven Crossover for \texorpdfstring{\\Real-Coded Genetic Algorithm}{Real-Coded Genetic Algorithm}}

\author{Takumi Nakane}
\affiliation{%
  \institution{University of Fukui}
  \city{Fukui} 
  \country{Japan} 
}
\email{t-nakane@monju.fuis.u-fukui.ac.jp}

\author{Xuequan Lu}
\affiliation{%
  \institution{Deakin University}
  \city{Victoria} 
  \country{Australia} 
}
\email{xuequan.lu@deakin.edu.au}

\author{Chao Zhang}
\affiliation{%
  \institution{University of Fukui}
  \city{Fukui} 
  \country{Japan} 
}
\email{zhang@u-fukui.ac.jp}


\begin{abstract}
In evolutionary algorithms, genetic operators iteratively generate new offspring which constitute a potentially valuable set of search history. To boost the performance of crossover in real-coded genetic algorithm (RCGA), in this paper we propose to exploit the search history cached so far in an online style during the iteration. Specifically, survivor individuals over past few generations are collected and stored in the archive to form the search history. We introduce a simple yet effective crossover model driven by the search history (abbreviated as SHX). In particular, the search history is clustered and each cluster is assigned a score for SHX. In essence, the proposed SHX is a data-driven method which exploits the search history to perform offspring selection after the offspring generation. Since no additional fitness evaluations are needed, SHX is favorable for the tasks with limited budget or expensive fitness evaluations. 
We experimentally verify the effectiveness of SHX over 4 benchmark functions. Quantitative results show that our SHX can significantly enhance the performance of RCGA, in terms of accuracy.
\end{abstract}

%
%



\maketitle

\section{Introduction}
\label{sec:introduction}
The exploration process of evolutionary algorithms (EAs) conducts the offspring generation and survivor individuals selection alternately and iteratively. Because of the offspring generation, a large number of candidate solutions (i.e., individuals) are sampled, accompanied by corresponding fitness values, genetic information and genealogy information. Such accumulated search data constitutes search history which can be very informative and valuable for boosting the overall performance. Therefore, the way of exploiting the search history truly matters to enable a better solution for the population without increasing the fitness evaluations (FEs).

In Real-coded genetic algorithm (RCGA) one of the part of EAs, the main efforts for improving the performance have been focused on the development of the crossover techniques \cite{herrera2003taxonomy}. Given different mechanisms, crossover methods can differ from (1) parent selection, (2) offspring generation, and (3) offspring selection. These associate the exploration ability with exploitation ability, and the degree and balance between both abilities affect the performance largely \cite{vcrepinvsek2013exploration}.

In this paper, we attempt to introduce a crossover method that effectively exploits the history data. At first, an archive is defined to collect the survivor individuals over generations as the search history. Then, the stored individuals are clustered by k-means \cite{lloyd1982least}, and each cluster is assigned a score depending on the number of belonging individuals. At last, offspring is generated and selected according to the scores. The proposed crossover operator, named search history driven crossover (SHX), generates offspring by considering the cluster scores. Since SHX enables an offspring selection mechanism, any existing parent selection and offspring generation mechanisms can be easily integrated with it. We evaluate the effectiveness of the proposed method using 4 benchmark functions. Two conventional crossover operators are employed, and the results with/without SHX are compared. Apart from the above, two archive update methods are also analyzed.

\section{Overview}
\label{sec:overview}
The proposed method manages not only population $\pop$ but also an archive $\arc$, which preserves survivors, throughout the generation alternation. $\pop$ and $\arc$ are initialized by randomly placing individuals in the search space. The archive update process is conducted after the survivor selection. Survivor individuals $\sur$ of current generation are aggregated into both $\pop$ and $\arc$ of the next generation. Different from conventional RCGA, individuals generated from parents $\pare$ are regarded as offspring candidates $\cand$. The main purpose of SHX is to narrow down $\cand$ to offspring $\offs$ according to the statistics provided by $\score$. $\score$ is calculated from the clustering result of archive and immediately impacts the offspring selection.

SHX can adopt any existing crossover operators. To show the performance increase brought by SHX, we choose the widely applied BLX-$\alpha$ \cite{eshelman1993real} and SPX \cite{tsutsui1999multi} for the offspring generation and compare the results in Sec. \ref{sec:experiments}.

\begin{table*}[!t]
\caption{The mean fitness values and standard deviations of the final-generation-elite over 10 runs. The best results in the BLX group (2nd$\sim$4th columns) and SPX group (5th$\sim$7th columns) are bold.}
\label{tab:result}
\footnotesize
\begin{tabular}{crrr|rrr}
\toprule
& BLX & SH-BLX\_random & SH-BLX\_sequential & SPX & SH-SPX\_random & SH-SPX\_sequential \\
Name & Mean (Std Dev) & Mean (Std Dev) & Mean (Std Dev) & Mean (Std Dev) & Mean (Std Dev) & Mean (Std Dev) \\
\midrule
Sphere & 5.45E+00 (1.77E+00) & \textbf{4.20E+00} (1.51E+00) & 4.29E+00 (\textbf{1.40E+00}) & 5.06E-03 (2.52E-03) & 1.51E-03 (\textbf{3.75E-04}) & \textbf{8.75E-04} (4.46E-04) \\
Rosenbrock & 6.31E+04 (3.16E+04) & 4.12E+04 (2.71E+04) & \textbf{3.38E+04} (\textbf{2.16E+04}) & 1.96E+01 (4.29E+00) & 1.30E+01 (3.43E+00) & \textbf{1.12E+01} (\textbf{1.80E+00}) \\
Rastrigin & 4.74E+01 (\textbf{5.85E+00}) & \textbf{4.13E+01} (7.93E+00) & 4.43E+01 (6.21E+00) & 3.78E+01 (4.92E+00) & 1.11E+01 (\textbf{4.81E+00}) & \textbf{8.32E+00} (5.11E+00) \\
Ackley 1 & 9.85E+00 (\textbf{5.64E-01}) & 8.75E+00 (1.06E+00) & \textbf{8.49E+00} (1.53E+00) & 6.76E-01 (2.46E-01) & 3.75E-01 (1.24E-01) & \textbf{1.88E-01} (\textbf{6.54E-02}) \\
\bottomrule
\end{tabular}
\end{table*}

\section{Survivor Archive} 
\label{sec:archive}
Given that SHX is to maintain the historical statistics $\score$ while producing offspring for the next generation, the archive $\arc$ is designed to store $\sur$ over few past generations and extracts statistics $\score$. In particular, k-means is employed to cluster the individuals in $\arc$, and $\score$ is a normalized frequency histogram to show the proportion regarding size of each cluster to $\arc$. The statistics can then be maintained by probabilistically assigning newly generated candidates to each cluster according to $\score$. In other words, the clusters generated by k-means reflect search regions with potential local optima in the fitness landscape.

To keep the computational cost brought by k-means within an acceptable and constant range, the archive size is fixed. Therefore, two types of update methods are considered in this work: (1) randomly selecting individuals in $\arc$ and replacing them with $\sur$ (denoted by $random$); (2) replacing a part of ${\arc}$ with $\sur$ in the order in which the individuals of ${\arc}$ arrived (denoted  by $sequential$). The performance comparison between these two approaches are discussed in Sec. \ref{sec:experiments}.

The update of $\arc$ and calculation of $\score$ are executed after the survivor selection. The centroids of the clusters are updated according to the updated $\arc$ by $random$ or $sequential$ approaches, and each individual in $\arc$ is assigned with an updated cluster label. After that, the normalized frequency histogram $\score$ for each cluster is calculated for further usage in offspring selection. Note that the initial centroids of the clusters in the current generation are inherited from the previous generation, as most individuals in $\arc$ between two consecutive generations are the same. 

\section{Search History Driven Crossover (SHX)}
\label{sec:shx}
SHX randomly selects parents by following the strategy of existing crossover operators, and excessively generates candidate offspring $\cand$ for further offspring selection. 
The size of $\cand$ is sufficiently larger than $\offs$ because $\cand$ must ensure a sufficient number of individuals that can be assigned to each cluster in $\arc$. Offspring selection narrows down $\cand$ to $\offs$ based on roulette wheel selection. Each proportion of the wheel relates to each possible selection (i.e., clusters), and $\score$ is used to associate a probability of selection with each cluster in $\arc$. This can also be viewed as a procedure that SHX preferentially selects individuals in more ``promising'' regions. Besides, the statistics of the population (e.g., cluster size) can be maintained between two consecutive generations because the new generation is sampled based on the statistics of the history. Also, the diversity of $\offs$ can be preserved because each individual from $\cand$ has a probability to be assigned to $\arc$.

\section{Experimental Results}
\label{sec:experiments}
The performance of SHX is investigated over 4 benchmark functions with 10 dimension settings. We comprehensively compare the performance of RCGA with/without SHX, and SHX is run with different settings of archive update methods ($random$/$sequential$) and offspring generation methods (BLX/SPX). All experiments are executed 10 times with different random seeds. We perform 100 generation alternation for a population composed of 100 individuals. SHX narrows down 180 candidates to 60 offspring. The archive stores survivors over 30 generations, and the survivors are further grouped into clusters. The number of clusters is half the number of the archive size.

The results of mean fitness values of final-generation-elite and standard deviations with respect to all combinations of functions and methods are displayed in Tab. \ref{tab:result}. We can observe the clear improvement of performance brought by SHX for both BLX and SPX. Since SHX manages an archive that stores search history over few generations, it can preserve some useful statistics (e.g., centroids of clusters), which can help to enhance BLX and SPX. On the other hand, SHX with sequential archive update achieves the best performance. One possible reason for $sequential$ outperforming $random$ in most cases is that $sequential$ removes the oldest individual which arrived first, and therefore SHX can select offspring according to the up-to-date search history to reflect the trend of evolution more sensitively.

\section{Conclusions}
\label{sec:conclusions}
In this paper, we have proposed a novel crossover model (SHX) which is simple yet effective and efficient. The key idea is to exploit search history over generations to gain useful information for generating offspring. Experimental results demonstrate that our SHX can significantly boost the performance of existing crossovers. As the future work, we would like to address parallelization to speed up SHX and adaptively setting of additional hyperparameters.

\bibliographystyle{ACM-Reference-Format}
\bibliography{bibliography} 

\end{document}